\documentclass[10pt, a4paper]{article}

\usepackage[final]{lrec2026} % this is the new style
\usepackage{multirow}
\usepackage{tabularx}
\usepackage{booktabs}
\usepackage{makecell}

\title{LombardoGraphia: Automatic Classification of Lombard Orthography Variants}

\name{Edoardo Signoroni, Pavel Rychlý} 

\address{Faculty of Informatics \\
Masaryk University \\ 
Botanická 68a, 602 00 Brno, Czechia \\
\texttt{e.signoroni@mail.muni.cz, pary@fi.muni.cz}}

\abstract{
Lombard, an underresourced language variety spoken by approximately 3.8 million people in Northern Italy and Southern Switzerland, lacks a unified orthographic standard. Multiple orthographic systems exist, creating challenges for NLP resource development and model training. This paper presents the first study of automatic Lombard orthography classification and LombardoGraphia, a curated corpus of 11,186 Lombard Wikipedia samples tagged across 9 orthographic variants, and models for automatic orthography classification. We curate the dataset, processing and filtering raw Wikipedia content to ensure text suitable for orthographic analysis. We train 24 traditional and neural classification models with various features and encoding levels. Our best models achieve 96.06\% and 85.78\% overall and average class accuracy, though performance on minority classes remains challenging due to data imbalance. Our work provides crucial infrastructure for building variety-aware NLP resources for Lombard.
\\ \newline \Keywords{Lombard, Low-Resource Languages, Orthography Classification, Language Identification, Italian Language Varieties} }

\begin{document}

\maketitleabstract

\section{Introduction}

%Low resource langs definition and motivation, digital divide etc
Recent advances in Large Language Models (LLMs) have significantly improved multilingual natural language processing, enabling high-quality machine translation (MT) and other downstream tasks for many major world languages. However, the benefits of these models are unevenly distributed, with underresourced languages being left behind in both training data and model capabilities.

Italy's uniquely diverse linguistic landscape features numerous regional varieties alongside Standard Italian, which only achieved widespread adoption following the birth of mass media. Consequently, most local varieties have declined due to marginalization and social stigma. Today, over 30 Italian language varieties are endangered \cite{UnescoAtlas}, though scattered interest is re-emerging in NLP.

One such language is Lombard, spoken in and around the Northern Italian region of Lombardy and in parts of Switzerland. Despite having an estimated number of speakers of 3.8 million people, NLP work on Lombard and its varieties is scarce \cite{ramponi-2024-language}, barring its use in an increasingly digitalized and interconnected world. However, developing tools and resources for Lombard is not a trivial task. As a primarily spoken language existing across a continuum of intercomprehensible varieties, it lacks a common standard orthography.

While some data is already available online, a fundamental first step is to correctly recognize and classify which of the many proposed orthographies is being used in a given text, with the ultimate aim to create a substantial corpus for each variety, and for Lombard as a whole. This would enable effective and language-aware training and developing NLP models and tools for the benefit of the Lombard community, without forcing an artificial standard on the speakers or unknowingly mixing together data for different varieties, potentially hampering downstream applications. This work addresses this gap by curating both the corpus and classification models needed for variety-aware Lombard NLP.

This paper introduces LombardoGraphia (lmo\_graphia)\footnote{The corpus, trained models, and code available on-line at \url{https://github.com/edoardosignoroni/lmo_graphia}}, a curated multi-orthography corpus, and presents the first study of automatic Lombard orthography classification. Our contributions are:

\begin{itemize}
    \item The \textbf{LombardoGraphia Corpus}: A curated dataset of 11,186 Lombard Wikipedia samples tagged across 9 orthographic variants (MILCLASS, LOCC, LORUNIF, SL, NOL, CRES, BREMOD, BERGDUC, LSI), with train/validation/test splits suitable for supervised learning. The corpus includes detailed metadata about orthographic systems and geographic distributions.
    
    \item \textbf{Trained models for automatic Lombard orthography classification}: Training and evaluation of 24 classification models featuring both traditional and neural approaches with different feature and encoding levels, enabling variety-aware NLP development for Lombard, supporting applications like corpus building, language identification, and orthographic normalization, while establishing benchmarks for future work.
\end{itemize}

\section{Lombard and its Orthographies}

\subsection{The Lombard Language}

Lombard is a language variety\footnote{In Italy, local language varieties are often stigmatized as a sign of ignorance and lack of integration, and denoted with the negatively-charged and linguistically improper connotation of \textit{dialetti} (dialects), implying a derivative status as "dialects of Standard Italian". As \citet{ramponi-2024-language} points out, the term \textit{language varieties} is a more neutral denomination, preventing judgment on the prestige and status of each language.} spoken in and around the Northern Italian region of Lombardy and in the Swiss Cantons of Ticino and Grisons\footnote{Lombard, in its Bergamasque variant, is spoken also in parts of southern Brazil, being brought there by immigrants during the 19th and 20th centuries \cite{paganessi-2017-berg}} by about 3.8 million people, where it exists alongside the official language in a state of \textit{dilalìa} \cite{ramponi-2024-language,berruto_dilalia}. Italian is used in all formal and official settings, whereas the local variety is more and more confined to informal situations, overlapping with Italian even in these domains.\footnote{In Switzerland the status of the local Lombard varieties is more vital, due to the better attitudes towards the language, with institutions such as the \textit{CDE – Centro di dialettologia e di etnografia} ("Centre for Dialectology and Ethnography")\footnote{\url{https://www4.ti.ch/decs/dcsu/cde/cde}} of Bellinzona doing research and maintain resources for the local Lombard variety.} For this reason, even if some historical literary traditions exist, Lombard varieties are primarily used in spoken and informal settings, and lack a codified written form, with the speakers improvising writing "as the words sound". When the speakers write their variety, it is often in code-switching. Lombard is promoted at a regional level in Lombardy by the Regional Law 25/2016.\footnote{\url{https://normelombardia.consiglio.regione.lombardia.it/normelombardia/accessibile/main.aspx?view=showpart\&idparte=lr002016100700025ar0024a}} 

It belongs to the Gallo-Romance-Cisalpine group of the Western Romance family of the Indo-European languages, and it is said to have between two and four varieties, the main ones being Western (in the provinces of Varese, Como, Lecco, Sondrio, Milan, Monza, Pavia and Lodi, in addition to Novara and Verbania in Piedmont and Canton Ticino in Switzerland) and Eastern Lombard (in the provinces of Bergamo, Brescia and Northern Cremona). These varieties, even with some phonetic, lexical, and grammatical differences, can be loosely considered to be one language continuum, since they are mutually intelligible \cite{ColuzziPlanning,BonfadiniDialetti,LoporcaroProfilo,ColuzziCase}. At the present day, the language is mostly used in oral conversation and no unified orthography exists.

\subsection{Lombard Orthographies}

Lombard, as a primarily spoken language, does not have a written standard common to all speakers, with the majority of them either avoiding writing or using their own subjective variants. Several orthographies were proposed, such the ones used on the Lombard Wikipedia,\footnote{\url{https://lmo.wikipedia.org/wiki/Wikipedia:GrafCat}} which can be divided in Pan-Lombard\footnote{These are also usually polynomic, that is their structure allows for local variations to spelling, while maintaining the same pronunciation rules for all Lombard dialects.}, Macro-dialectal, and Local Orthographies. Below, we will introduce the ones that are relevant for the Wikipedia dataset and this work.

\subsubsection{Pan-Lombard Orthographies}

\textbf{Noeuva Ortografia Lombarda} The \textit{Noeuva Ortografia Lombarda} (NOL, "New Lombard Orthography")\footnote{\url{https://lmo.wikipedia.org/wiki/Noeuva_Ortografia_Lombarda}; \url{https://academiabonvesin.eu/noeuva-ortografia-lombarda/}} is an orthography based on the writing tradition of different Lombard variants and created for the whole Lombard language. It is a polynomic system which aims to give all Lombard speakers the flexibility to write with the same rules, while allowing to keep and show local differences. It is also easy to write with a standard Italian keyboard, avoiding symbols and diacritics borrowed from foreign languages, such as the German \textit{umlaut}.

\textbf{Scriver Lombard} \textit{Scriver Lombard} (SL, "Writing Lombard")\footnote{\url{https://lmo.wikipedia.org/wiki/Scriver_Lombard}; \url{http://inlombard.eu5.net/indexLmo.html}} is another polynomic orthography proposed by \citet{BrascaScriver}. This system was inspired by older Lombard literary tradition, such as the medieval author Bonvesin de la Riva (\~1250-1313/15) and others. It proposes a "partially uniform" system for all speakers that still allows for minimal variations. The author states that SL has the revitalization and intergenerational transmission of Lombard as its ultimate objective.

\subsubsection{Macro-Dialectal Orthographies}

\textbf{Urtugrafia Insübrica Ünificada} The \textit{Urtugrafia insübrica ünificada} (LOCC, "Unified Insubric Orthography") \footnote{\href{https://lmo.wikipedia.org/wiki/Urtugrafia_ins\%C3\%BCbrica_\%C3\%BCnificada}{\texttt{https://lmo.wikipedia.org/wiki/ Urtugrafia\_insübrica\_ünificada}} \textit{Insubria} is an historical region that during the Classical antiquity was populated by the Insubres. For a time, it denoted the western part of Lombardy, plus Ticino and the province of Novara. Today, it can also be used to refer to Milan and the surrounding territories.} was used by the journal of the cultural association "\textit{La Vus de l'Insübria}" ("The Voice of Insubria") for the local Lombard variant. It is a phonemic system based on Italian graphemes with diacritics to distinguish between open and closed vowels, searching for a compromise between the different local Insubric pronunciation. 

\textbf{Ortograféa Orientàl Ünificàda} The \textit{Ortograféa orientàl ünificàda} (LORUNIF, "Unified Oriental Orthography")\footnote{\href{https://lmo.wikipedia.org/wiki/Ortograf\%C3\%A9a_orient\%C3\%A0l_\%C3\%BCnific\%C3\%A0da}{\texttt{https://lmo.wikipedia.org/wiki/ Ortograféa\_orientàl\_ünificàda}}} is an attempt to give one phonematic writing system to the most spoken variants of Eastern Lombard, \textit{Bergamàsch} ("Bergamasque"), \textit{Bresà} ("Brescian"), and \textit{Cremàsch} ("Cremasque"). It is based on several attested traditions, in particular the system used by Canossi (1862-1943) and Melchiori \cite{melchiori} for Brescian, and by Zappettini \cite{zappettini} and the cultural association "\textit{Dücàt de Piàsa Puntìda}" ("Duchy of Pontida Square") for Bergamasque. It thus keeps the same rules of the \textit{Ortografìa bresàna modèrna} ("Modern Brescian Orthography") and of the \textit{Ortografìa del Dücàt Semplificàda} ("Simplified Duchy Orthography") with minor variations.

\subsubsection{Local Orthographies}

\textbf{Ortografia Milanesa} The \textit{Ortografia Milanesa} (MILCLASS, "Milanese Orthography"),\footnote{\url{https://lmo.wikipedia.org/wiki/Ortografia_milanesa}} also called \textit{classega} ("Classical") is rooted in the tradition of the Milanese literature, starting from the XVII century with the works of Carlo Maria Maggi (1630-1699). This system appears as a compromise between the Italian orthography and the French or Provencal one. With the spread of Italian, the Milanese orthography started to be considered too complicated and became obsolete, thus being referred to as "\textit{classega}". The \textit{Circol Filògich Milanes} ("Milanese Philological Club") adapted the classical orthography for the modern use.

\textbf{Ortograféa del Dücat} The \textit{Ortograféa del Dücat} (BERGDUC, "Orthography of the Duchy")\footnote{\href{https://lmo.wikipedia.org/wiki/Ortograf\%C3\%A9a_del_D\%C3\%BCcat}{\texttt{https://lmo.wikipedia.org/wiki/ Ortograféa\_del\_Dücat}}} is the system used by the cultural association "\textit{Dücàt de Piàsa Puntìda}", founded in 1924. It is based on older systems from the XIX century, with the vocabularies from Zappettini \cite{zappettini} and Tiraboschi \cite{tiraboschi}. It was used also by the most important Bergamasque authors of the first half of the 1900.

\textbf{Grafia LSI}\footnote{\url{https://it.wikipedia.org/wiki/Ortografia_ticinese}. The tag LSI comes from "\textit{LSI - Lessico dialettale della Svizzera italiana}" ("Dialectal Lexicon of Italian Switzerland"), a vocabulary of the Lombard variants in Switzerland.} First used in 1907 to compile the \textit{Vocabolario dei dialetti della Svizzera italiana} ("Vocabulary of the Dialects of Italian Switzerland")\footnote{\href{https://www4.ti.ch/decs/dcsu/cde/pubblicazioni/vocabolario-dei-dialetti-della-svizzera-italiana}{\texttt{https://www4.ti.ch/decs/dcsu/cde/ pubblicazioni/vocabolario-dei-dialetti-della-svizzera-italiana}}}, it is based on the Italian orthography with the addition of umlauts and elements from Classical Milanese. It is maintained and developed by the \textit{Centro di dialettologia e di etnografia} ("Centre for Dialectology and Ethnography") of Canton Ticino, where it is primarily used (together with Grigioni) for street signs and local toponyms. This makes it the only Lombard orthography with an "official" status. It is a phonologic system created for the classification and study of the different variants of Lombard in Switzerland.

\textbf{Ortografìa Bresàna Moderna} The \textit{Ortografìa Bresàna Moderna} (BREMOD, "Modern Brescian Orthography")\footnote{\href{https://lmo.wikipedia.org/wiki/Ortograf\%C3\%ACa_del_Bres\%C3\%A0}{\texttt{https://lmo.wikipedia.org/wiki/ Ortografìa\_del\_Bresà}}} is based on the system used by Angelo Maria Canossi (the most important contemporary Brescian author), even if with some variations to rationalize and solve some ambiguities. It largely uses the same rules as Italian, but employs diacritics such as accents and umlauts to distinguish between different vowel sounds and mark the tonic accent.

\textbf{Urtugrafìa Cremàsca} The "Urtugrafìa Cremàsca" (CRES, "Cremasque Orthography")\footnote{\href{https://lmo.wikipedia.org/wiki/Urtugraf\%C3\%ACa_Crem\%C3\%A0sca}{\texttt{https://lmo.wikipedia.org/wiki/ Urtugrafìa\_Cremàsca}}} is similar to other Eastern Lombard systems in that it is largely based on Italian rules with some added diacritics to mark accents and roundedness and openness.

%TABLE OF COMPARISON OF ORTHOGRAPHIES
\begin{table*}
\centering
\begin{tabularx}{\textwidth}{lX}
    \toprule
    \textbf{Orthography} & \textbf{Sample Text} \\
    \midrule
     \textbf{MILCLASS} & \textit{Sera settada in terra col coo in man, e i gombed sui genœucc: me ziffolava el vent in di cavij: demaneman che vegneva on quaj bôff, el me portava comè ona vôs che vegneva de lontan;} [...] \newline "I was seating on the ground with the head in the hands, and the elbows on the knees: as it came in whiffs, it brought me as a voice from afar; [...]" \\
     \midrule
     \textbf{LOCC} & \textit{Sera setada in tera cul coo in man, e i gumbed süi genögg: me zifulava el vent in di cavii: demaneman che vegneva un quai buf, el me purtava cumè una vus che vegneva de luntan;\textbf{}} [...]  \\
     \midrule
     \textbf{SL} & \textit{S’era setada in terra col coo in man, e i gombeds sui jenœgg: me cifolava el vent ind i cavei: demaneman qe vegneva un quai bof, al me portava comè una vox qe vegneva de lontan;} [...] \\
     \midrule
     \textbf{NOL} & \textit{S’era setada in terra col coo in man, e i gombet sui sgenoeugg: me scifolava el vent in di cavej: demaneman che vegneva un quaj bof, el me portava comè una vos che vegneva de lontan;} [...]  \\
     \midrule
     \textbf{LSI} & \textit{Cula paròla Ticines a sa inteend i dialett che i è parlaa in Tesin (vün dai 26 cantón svizzer) e in Mesulcina e Calanca (dó vall dal Canton Grison).} [...] \newline "With the word \textit{Ticines}, we mean the dialects spoken in Ticino (one of the 26 Swiss cantons) and in Mesolcina and Calanca (two valleys of Grisons)" \\
     \midrule
     \textbf{LORUNIF} & \textit{L'ortografìa orientàl ünificàda l'è 'n tentatìf de dàga 'na rispòsta al bizògn de 'n sistéma de scritüra bù per töte le varietà piö parlàde del Lombàrt Orientàl} [...] \newline "The unified oriental orthography is an attempt to answer the need for a writing system for all the most spoken varieties of Eastern Lombard." \\
     \midrule
     \textbf{BREMOD} & \textit{Le régole dopràde endèi artìcoi marcàcc come scriìcc segónt l'ortografìa modèrna le se rifà a la grafìa del Canossi che l'è sènsa döbe l'autùr dialetàl bresà contemporàneo piö 'mportànte.} \newline "The rules used in the articles tagged as written following the modern orthography are inspired by the orthography of Canossi who without a doubt was the most important contemporary brescian dialectal author." \\
     \midrule
     \textbf{BERGDUC} & \textit{L'ortograféa del Dücàt l'è ol sistéma de régole de trascrissiù del dialèt bergamasch dovrade de l'associassiù del Dücàt de Piassa Püntida in di sò püblegassiù.} \newline "The orthography of the Duchy is the system of transcription rules of the Bergamasque dialect used by the association of the Duchy of Piazza Pontida in its publications." \\
     \midrule
     \textbf{CRES} & \textit{Per capì la scritüra cremàsca sèrf adóma poche régule, simìi a chèle per scrìf an italià. [...] I acént i è quasi sèmper segnàt, anche 'ndù i è mia stretamént necesàre.} \newline "To understand Cremasque writing only few rules are needed, similar to those for writing in Italian. [...] The accents are almost always marked, even where they are not strictly needed." \\
     \bottomrule
\end{tabularx}
\caption{Summary of the Lombard orthographies in the dataset. The first column give their abbreviated name, while the second shows a text sample for each one. When available (MILCLASS, LOCC, SL, NOL) we used the same text.}{\footnotemark}
\label{tab:graf_sample}
\end{table*}

\footnotetext{From \textit{La Fuggitiva} ("The Fugitive") by Tommaso Grossi (1791-1853). Retrieved from \url{https://lmo.wikipedia.org/wiki/Dialett_milanes}. Own translation.}

Table \ref{tab:graf_sample} gives samples of text in each orthography.
The Pan-Lombard orthographies (NOL and SL) are explicitly built to be used independently by all Lombard speakers, regardless of the variant. The vast majority of the data, however, is written either in macro-dialectal or local orthographies which are by nature intertwined with the Lombard variants for which they are constructed, and used by the speakers of those specific varieties. Thus, orthography and variant are closely connected in practice, leading not only to differences in orthographical rules, but also in lexical choices and topics. So, while at a first glance many orthographies look similar from the rules point-of-view, it is still important to distinguish between them in order to grasp subtle differences and wider linguistic implications.

\section{Related Work}

To our knowledge, no prior published work was done for the automatic classification of Lombard orthographies. In the following section, we thus briefly introduce Language Identification (LI) for similar variants (Subsection \ref{sec:related_li}) and some other NLP work and resources for Lombard (Subsection \ref{sec:related_lmo}).

\subsection{LI for Similar Variants} \label{sec:related_li}

This work can be framed as a language identification task, that is the problem of determining the natural language or variety that a document is written in. While one line of research aim at broadening the number of languages supported by one single system, another one is focused on groups of similar languages, covering idioms from various groups and families. Training a system to discern between similar languages, dialects, or variants is harder than having it distinguish completely different languages. For a broader survey of LI, we refer the reader to \citet{Jauhiainen-etal-2019-survey}. 

The VarDial evaluation campaigns have long served as the primary benchmark for similar language and dialect identification \cite{zampieri-etal-2020-natural}. Recently, the campaign explicitly addressed the linguistic landscape of Italy through the Identification of Languages and Dialects of Italy shared task introduced in VarDial 2022 \cite{aepli-etal-2022-findings}. Participants were tasked with classifying text across several Italian language varieties, Lombard included. Findings from the shared task demonstrated that traditional machine learning models and simple character-level neural networks frequently outperformed massive pre-trained language models on this specific problem \cite{ceolin-2022-neural}. This reflects our own findings for Lombard orthographies, where classical ML approaches prove highly competitive and more robust.

Character \textit{n}-grams have been used effectively in text categorization and LI for decades \cite{Jauhiainen-etal-2019-survey}, since the milestone method of \citet{cavnar-trenkle-1994-ngram} and its off-the-shelf implementation of \citet{van1997textcat}. 

\subsection{NLP for Lombard} \label{sec:related_lmo}

Beyond pure language identification, there is a growing recognition of the need for speaker-centric, variety-aware NLP for Italy's endangered languages \cite{ramponi-2024-language}. Recent efforts have begun to map the diatopic variation of Italy in digital spaces, such as social media corpora \citelanguageresource{ramponi-casula-2023-diatopit}. However, most existing computational approaches still implicitly treat local languages as unified, monolithic entities with standardized writing systems. By framing orthography classification as a foundational precursor to dialect identification, we address the reality that varieties like Lombard are a linguistic continuum lack a single codified standard, a practical dimension often overlooked in broad-coverage language identification systems.

While at least some research has been done on other language varieties of Italy, NLP work explicitly for Lombard is scarce \cite{ramponi-2024-language}. 

\citetlanguageresource{signoroni2022piotost} describes a human-evaluated, revised, and corrected Lombard-Italian parallel corpus destined to train machine translation systems. With the help of bilingual annotators, they audit an automatically aligned Wikipedia corpus from OPUS \citelanguageresource{opus}.

Usually if Lombard is present, it is in a multilingual setting: it is part of benchmark datasets (FLORES+ and OLDI Seed) \citelanguageresource{Costa-jussa2024}, supported by some multilingual MT models (NLLB-200) \cite{nllbteam2022languageleftbehindscaling} and LLMs (mBERT) \cite{devlin-etal-2019-bert}. 

\section{Methodology}

\begin{table*}[htb!]
\centering
\tiny
\setlength{\tabcolsep}{2pt}
\resizebox{\textwidth}{!}{
\begin{tabular}{lrrrrrrrrrrrr}
\toprule
\multirow{2}{*}{\textbf{Class}} & \multirow{2}{*}{\textbf{Type}} & \multirow{2}{*}{\textbf{Total Lines}} & \multicolumn{2}{c}{\textbf{Train}} & \multicolumn{2}{c}{\textbf{Valid}} & \multicolumn{2}{c}{\textbf{Test}} & \multicolumn{2}{c}{\textbf{Dataset Total}} & \multicolumn{2}{c}{\textbf{Removed}} \\
\cline{4-13}
& & & \textbf{N} & \textbf{\%} & \textbf{N} & \textbf{\%} & \textbf{N} & \textbf{\%} & \textbf{N} & \textbf{\%} & \textbf{N} & \textbf{\%} \\
\midrule
MILCLASS & Local & 79,196 & 3,606 & 40.29 & 471 & 42.13 & 446 & 39.89 & 4,523 & 40.43 & 74,673 & 94.29 \\
LOCC & Macro-dialectal & 34,794 & 2,907 & 32.48 & 345 & 30.86 & 380 & 33.99 & 3,632 & 32.47 & 31,162 & 89.56 \\
LORUNIF & Macro-dialectal & 76,455 & 1,901 & 21.24 & 240 & 21.47 & 229 & 20.48 & 2,370 & 21.19 & 74,085 & 96.90 \\
SL & Pan-Lombard & 3,761 & 174 & 1.94 & 22 & 1.97 & 16 & 1.43 & 212 & 1.90 & 3,549 & 94.36 \\
NOL & Pan-Lombard & 2,349 & 109 & 1.22 & 9 & 0.81 & 8 & 0.72 & 126 & 1.13 & 2,223 & 94.64 \\
CRES & Local & 633 & 98 & 1.09 & 17 & 1.52 & 19 & 1.70 & 134 & 1.20 & 499 & 78.83 \\
BREMOD & Local & 2,675 & 94 & 1.05 & 9 & 0.81 & 13 & 1.16 & 116 & 1.04 & 2,559 & 95.66 \\
BERGDUC & Local & 990 & 59 & 0.66 & 5 & 0.45 & 6 & 0.54 & 70 & 0.63 & 920 & 92.93 \\
LSI & Local & 6 & 2 & 0.02 & 0 & 0.00 & 1 & 0.09 & 3 & 0.03 & 3 & 50.00 \\
\midrule
no\_tag & - & 94,520 & - & - & - & - & - & - & - & - & 94,520 & 100.00 \\
\midrule
\textbf{Total} & & \textbf{295,379} & \textbf{8,950} & \textbf{100.0} & \textbf{1,118} & \textbf{100.0} & \textbf{1,118} & \textbf{100.0} & \textbf{11,186} & \textbf{100.0} & \textbf{284,193} & \textbf{96.21} \\
\bottomrule
\end{tabular}}
\caption{Distribution of orthographic classes across train, validation, and test sets. \textbf{Total Lines} shows the number of lines in the Wikipedia corpus before filtering. \textbf{Dataset Total} reports the number of lines in the cleaned corpus. \textbf{Removed} shows lines filtered out due to quality criteria (N) and the percentage removed relative to \textbf{Total Lines} (\%). The no\_tag category contains lines without orthography tags that were excluded from the dataset.}
\label{tab:dataset_distribution}
\end{table*}

\subsection{The LombardoGraphia Corpus}

\textbf{Data Source and Collection}
We collect data and text from the Lombard Wikipedia.\footnote{\url{https://lmo.wikipedia.org/wiki/Pagina_principala} We retrieved the dumps from July 2, 2025.} The Lombard Wikipedia suggests writing the articles in one of the pan-Lombard orthographies, the SL and NOL, but accepts also other macro-dialectal and local systems. It is strongly suggested to mark which variant is used in an article by using the corresponding template. This feature was crucial to build and process the data.

\textbf{Filtering} We subject the data to extensive filtering.

We first process the raw Wikipedia XML using \texttt{wikiextractor}\footnote{\url{https://github.com/attardi/wikiextractor}} to remove markup and extract plain text. 

We scan each entry to find if there is an orthography tag; which we use to categorize the article. We assume these as gold data, since they are chosen by the authors of the relative article. The rest of the articles are assigned to the "no\_tag" class.
We extract 295,379 total lines from the Wikipedia dump, of which 200,859 contain orthographic tags and 94,520 are untagged.

We deduplicate the resulting lines, and then we manually check the output to further remove other noise: leftover boilerplate text, sentences in a language other than Lombard,\footnote{Mostly Italian, English, and some text in Cyrillic alphabet.} short lines\footnote{Lines with 3 words or less were mostly foreign named entities, or dates.}, recurring lines of bot-generated articles,\footnote{Such as pages about years, towns, and stations, etc., e.g. \textit{El 1901 a l'è 'n ann del secol quell de vint.} ("1901 is a year of the XX century.")} and the like. 

This ensures that the text in the corpus is primarily in Lombard, contains substantive linguistic content, exhibits orthographic features, and has clear orthographic tags assigned by human contributors.

\textbf{Corpus Description}
Table~\ref{tab:dataset_distribution} presents the corpus composition. LombardoGraphia contains 11,186 samples distributed across 9 orthographic classes, with imbalance reflecting the orthographic preferences of Wikipedia contributors. 
Major classes represent 94.01\% of the data (MILCLASS, 40.43\%; LOCC, 32.47\%; LORUNIF, 21.19\%), while minor classes are 5.96\% of the examples (SL, 1.90\%; NOL, 1.13\%; CRES, 1.20\%; BREMOD, 1.04\%; BERGDUC, 0.63\%). LSI has only a minimal 0.03\%, with only 3 samples.

The corpus is split into training (8,950 samples, 80\%), validation (1,118 samples, 10\%), and test (1,118 samples, 10\%) sets. The class distribution is similar across all sets.

The vast majority of the training data is in Western Lombard (MILCLASS+LOCC, 72.77\%), around a quarter of the lines is in Eastern Lombard (LORUNIF+CRES+BREMOD+BERGDUC, 24.04\%), while just a tiny fraction is written in a Pan-Lombard orthography (SL+NOL, 3.16\%), even if the use of SL and NOL is clearly suggested by Wikipedia. 

After cleaning, 96.21\% of lines were filtered out. LSI was left with only 2 viable examples in the train split, thus we decided to exclude it from the experiments.

LombardoGraphia is released in JSONL format:
\begin{verbatim}
{"text": "sample text in Lombard", 
"tag": "ORTHOGRAPHY_CLASS"}
\end{verbatim}

\subsection{Model Training}

We train both traditional and neural classifiers on the tagged Wikipedia corpus. We train traditional models using byte, char, or word 1- to 4-grams; and a combination of all three features. Class imbalance is addressed through balanced class weighting in all classifiers. We evaluate the models on the validation set during training and choose the one with the best accuracy score. Final evaluation is performed on the held-out test set. We train 24 models in total, combining four traditional classifiers (Logistic Regression, SVM, Naive Bayes, Random Forest) and four neural architectures (LSTM, CNN, Deep CNN, Transformer) with four feature types for traditional models (character n-grams, byte n-grams, word n-grams and combined features), and two encodings for neural ones (byte and characters).

\subsection{Traditional Approaches}

We employ four traditional machine learning classifiers using \texttt{scikit-learn} implementations. Models use TF-IDF weighted byte, character, or word n-grams (n=1-4) with a maximum of 10,000 features per vectorizer. The byte and character-level approach captures spelling patterns without requiring explicit tokenization, while words are trivially split as white-spaced strings. Multiple feature types (character, byte, and word n-grams) can be combined by concatenating their feature vectors. 

\textbf{Logistic Regression}
We use multinomial logistic regression with L-BFGS optimization (max\_iter=1000) and balanced class weights. The multinomial formulation enables direct multi-class classification across all orthographic variants simultaneously.

\textbf{Support Vector Machine}
We employ LinearSVC with balanced class weighting (max\_iter=4000). The linear kernel provides efficient training while maintaining good performance on the high-dimensional sparse feature space.

\textbf{Naive Bayes}
We apply MultinomialNB with default scikit-learn parameters. Despite its independence assumption, Naive Bayes often performs well on text classification with sparse features.

\textbf{Random Forest}
We train an ensemble of 100 decision trees with balanced class weights. The ensemble approach helps capture complex patterns in orthographic variation while being robust to overfitting.

\subsection{Neural Approaches}

We train neural classifiers with both character-level and byte-level encoding. Character encoding uses a learned vocabulary (with special tokens for padding and unknown characters), while byte encoding uses fixed representations (0-255 plus padding at 256). Models operate on sequences of maximum length 200, with embedding dimension 128, trained using the Adam optimizer (learning rate 0.001) and batch size 128 for 10 epochs. Best models are selected based on validation accuracy and tested on the held out test set. All neural models use cross-entropy loss and implement early stopping based on validation accuracy to prevent overfitting.

\textbf{Long Short-Term Memory}
Our bidirectional LSTM uses 256 hidden units per direction across 2 layers with 0.5 dropout. The final classification uses the concatenated last hidden states from both directions, allowing the model to capture both left and right context for each character or byte.

\textbf{Convolutional Neural Network}
We implement two CNN architectures. The \textit{Wide CNN} uses 3 parallel convolutional layers with different kernel sizes (3, 4, and 5) and 256 filters each, followed by max pooling and concatenation. This architecture captures different n-gram patterns simultaneously. The \textit{Deep CNN} employs 6 stacked convolutional layers (256 filters each, alternating kernel sizes of 7 and 3) with batch normalization and progressive max pooling (stride 2).

\textbf{Transformer}
Our transformer uses 4 layers with 8 attention heads and learned positional encoding. Dropout is set to 0.3. We apply mean pooling over non-padding tokens before classification, allowing the model to attend to relevant patterns regardless of position in the sequence.

\begin{table*}[htb!]
\centering
\tiny
\setlength{\tabcolsep}{2.5pt}
\resizebox{\textwidth}{!}{%
\begin{tabular}{l|cccccccc|cc}
\hline
\textbf{Model} & \textbf{MILCLASS} & \textbf{LOCC} & \textbf{LORUNIF} & \textbf{SL} & \textbf{NOL} & \textbf{CRES} & \textbf{BREMOD} & \textbf{BERGDUC} & \textbf{Overall} & \textbf{Avg Class} \\
\hline
\multicolumn{11}{l}{\textit{Traditional Models - Logistic Regression}} \\
\hline
Log. byte & 97.31 & 92.89 & 89.08 & \textbf{100.0} & \textbf{75.00} & 94.74 & 53.85 & \textbf{83.33} & 93.38 & \textbf{85.78} \\
Log. byte+char+word & 97.31 & 94.74 & 94.76 & \textbf{100.0} & \textbf{75.00} & \textbf{100.0} & 46.15 & 66.67 & 95.08 & 84.33 \\
Log. char & 95.07 & 93.42 & 84.28 & \textbf{100.0} & \textbf{75.00} & 94.74 & 53.85 & \textbf{83.33} & 91.67 & 84.96 \\
Log. word & 94.62 & 94.21 & 88.65 & 93.75 & 37.50 & 94.74 & \textbf{69.23} & 66.67 & 92.39 & 79.92 \\
\hline
\multicolumn{11}{l}{\textit{Traditional Models - Support Vector Machine}} \\
\hline
SVM byte & 97.76 & 94.21 & \textbf{99.13} & 87.50 & \textbf{75.00} & 94.74 & 30.77 & 50.00 & 95.43 & 78.64 \\
SVM byte+char+word & 97.76 & \textbf{96.58} & \textbf{99.13} & 93.75 & 50.00 & 94.74 & 23.08 & 50.00 & \textbf{96.06} & 75.63 \\
SVM char & 98.21 & 94.74 & 98.25 & 93.75 & 50.00 & 94.74 & 30.77 & 50.00 & 95.52 & 76.31 \\
SVM word & 95.52 & 95.53 & 95.20 & 93.75 & 12.50 & 89.47 & 30.77 & 50.00 & 93.73 & 70.34 \\
\hline
\multicolumn{11}{l}{\textit{Traditional Models - Naive Bayes}} \\
\hline
NB byte & \textbf{98.88} & 91.05 & 93.45 & 0.00 & 0.00 & 0.00 & 0.00 & 0.00 & 89.62 & 35.42 \\
NB byte+char+word & 98.21 & 93.42 & 96.07 & 0.00 & 0.00 & 0.00 & 0.00 & 0.00 & 90.69 & 35.96 \\
NB char & 98.65 & 91.05 & 94.32 & 0.00 & 0.00 & 0.00 & 0.00 & 0.00 & 89.70 & 35.50 \\
NB word & 97.09 & 92.37 & 94.76 & 12.50 & 0.00 & 15.79 & 0.00 & 0.00 & 90.06 & 39.06 \\
\hline
\multicolumn{11}{l}{\textit{Traditional Models - Random Forest}} \\
\hline
RF byte & 98.43 & 92.37 & 96.94 & 68.75 & 0.00 & 68.42 & 0.00 & 0.00 & 92.75 & 53.11 \\
RF byte+char+word & 98.65 & 92.11 & 95.63 & 75.00 & 0.00 & 68.42 & 7.69 & 0.00 & 92.66 & 54.69 \\
RF char & 98.65 & 93.16 & 95.20 & 50.00 & 0.00 & 68.42 & 7.69 & 0.00 & 92.57 & 51.64 \\
RF word & 97.31 & 87.89 & 93.89 & 68.75 & 0.00 & 63.16 & 15.38 & 0.00 & 90.24 & 53.30 \\
\hline
\multicolumn{11}{l}{\textit{Neural Models}} \\
\hline
CNN byte & 97.98 & 95.53 & 97.38 & 75.00 & 12.50 & 78.95 & 7.69 & 16.67 & 94.27 & 60.21 \\
CNN char & 98.21 & 94.47 & 98.69 & 87.50 & 12.50 & 63.16 & 7.69 & 16.67 & 94.18 & 59.86 \\
Deep CNN byte & 83.86 & 94.74 & 92.14 & 75.00 & 12.50 & 73.68 & 15.38 & 0.00 & 87.20 & 55.91 \\
Deep CNN char & 97.09 & 92.37 & 96.94 & 0.00 & 12.50 & 57.89 & 7.69 & 0.00 & 91.23 & 45.56 \\
LSTM byte & 97.53 & 91.05 & 95.63 & 0.00 & 0.00 & 63.16 & 0.00 & 0.00 & 90.60 & 43.42 \\
LSTM char & 98.21 & 91.05 & 95.63 & 0.00 & 0.00 & 68.42 & 0.00 & 16.67 & 91.05 & 46.25 \\
Transformer byte & 96.64 & 85.00 & 94.76 & 68.75 & 0.00 & 5.26 & 0.00 & 0.00 & 88.00 & 43.80 \\
Transformer char & 96.41 & 88.95 & 93.89 & 0.00 & 0.00 & 31.58 & 0.00 & 0.00 & 88.54 & 38.85 \\
\hline
\multicolumn{11}{l}{\textit{Performance Statistics}} \\
\hline
Best accuracy & \textbf{98.88} & \textbf{96.58} & \textbf{99.13} & \textbf{100.0} & \textbf{75.00} & \textbf{100.0} & \textbf{69.23} & \textbf{83.33} & \textbf{96.06} & \textbf{85.78} \\
Worst accuracy & 83.86 & 85.00 & 84.28 & 0.00 & 0.00 & 0.00 & 0.00 & 0.00 & 87.20 & 35.42 \\
Accuracy range & 15.02 & 11.58 & 14.85 & 100.0 & 75.00 & 100.0 & 69.23 & 83.33 & 8.86 & 50.36 \\
\hline
\end{tabular}
}
\caption{Complete per-class accuracy (\%) across all 24 models, with overall and average class accuracy. Bold indicates the best performance per metric. Classes are ordered by sample size. Overall accuracy measures performance on all test samples, while high average class accuracy shows consistent performance across classes.}
\label{tab:perclass_results}
\end{table*}

\begin{figure*}[htb!]
    \centering
    \includegraphics[width=\textwidth]{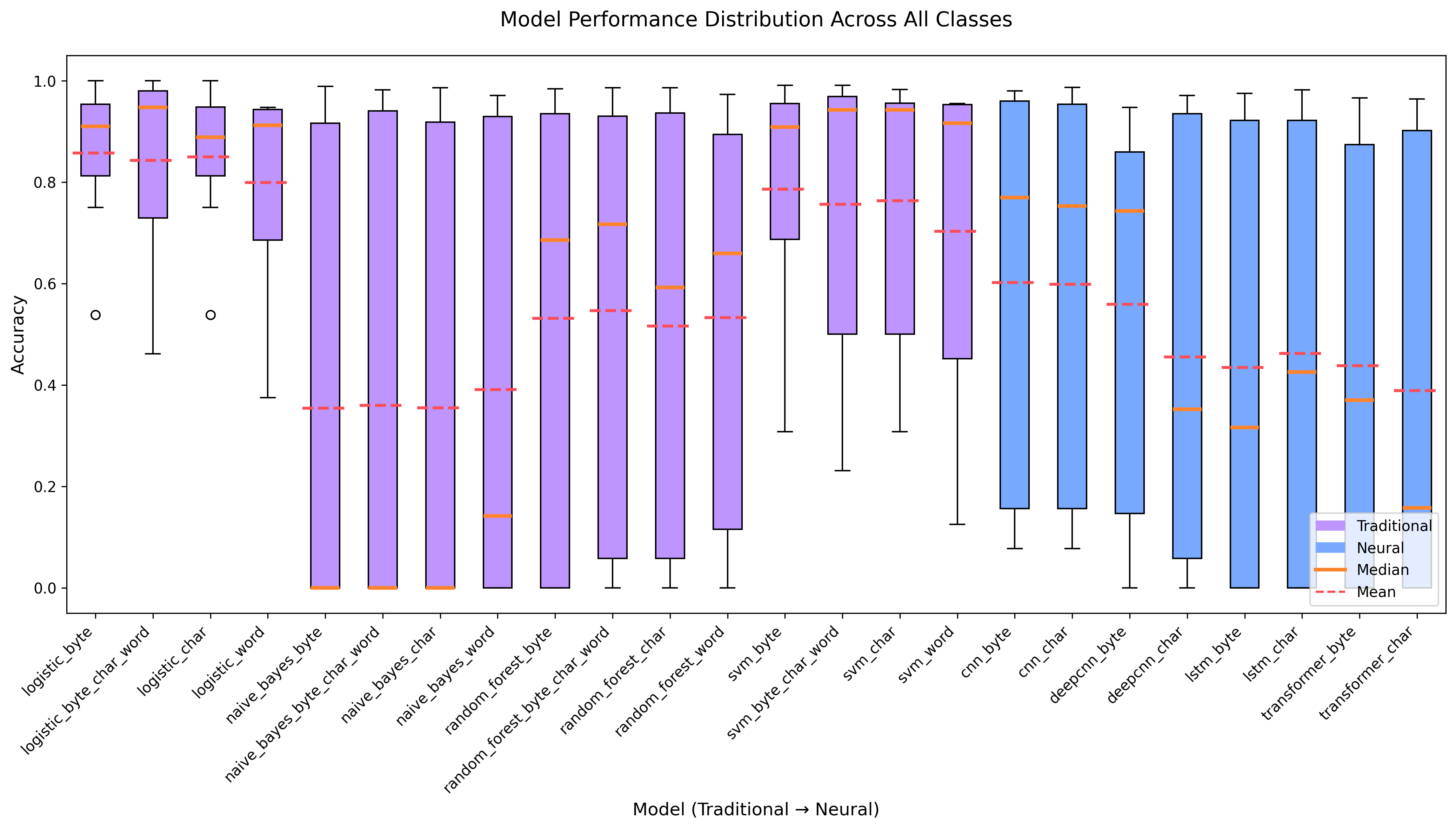}
    \caption{Boxplot of the results for all models. Traditional models are in purple, and neural models are in blue.}
    \label{fig:per-class-barplot}
\end{figure*}

\section{Results}

We evaluate all models on a held-out test set of 1,118 samples across 9 orthographic classes. Table \ref{tab:perclass_results} shows overall results and detailed per-class accuracies for selected top-performing models. Figure \ref{fig:per-class-barplot} visualizes the performance variation across all models.

\textbf{Overall Performance}
The SVM classifier with combined byte, character, and word n-gram features achieves the highest overall accuracy at 96.06\%, correctly classifying 1,074 out of 1,118 test samples. However, when considering average per-class accuracy, Logistic Regression with byte features performs best at 85.78\%, indicating more balanced performance across all orthographies. This is relevant for downstream application, given the dataset's severe class imbalance, where the three dominant classes (MILCLASS, LOCC, LORUNIF) represent 94.36\% of the test set.

Combined features (byte+char+word) generally improve overall accuracy for traditional models, with SVM and Logistic Regression both achieving their best overall performance using this combination. However, the same cannot be said for average class accuracy, which is worse or unchanged when the combined features are used. 

Naive Bayes is significantly worse than other traditional methods, due to its nearly complete inability to correctly classify the minority classes.  

Among neural models, the CNN architectures achieve the highest overall accuracy (94.27\% for byte-level, 94.18\% for character-level), approaching the performance of traditional methods. However, all neural architectures show substantially lower average class accuracy compared to Logistic Regression and SVM, since they unsurprisingly struggle with minority classes. The best neural model (CNN with byte encoding) achieves only 60.21\% average class accuracy, compared to 85.78\% for Logistic Regression with the same features.

\textbf{Per-Class Performance}
Performance varies across orthographic classes, as shown in Table~\ref{tab:perclass_results}. The three major classes (MILCLASS, LOCC, LORUNIF) achieve consistently high accuracy across nearly all models, with most configurations exceeding 90\%. The best performance on MILCLASS reaches 98.88\% (Naive Bayes with byte features), while LOCC peaks at 96.58\% (SVM with combined features) and LORUNIF at 99.13\% (both SVM variants with byte+char+word and byte features).

Minority classes show extreme performance variation and much lower overall accuracy. The accuracy range (difference between best and worst model) quantifies this variation: SL, NOL, CRES, BREMOD, and BERGDUC all show ranges of 0.69-1.0, meaning the best model achieves 69-100 percentage points higher accuracy than the worst. For these low-resource classes, Logistic Regression and SVM with byte or combined features consistently outperform neural methods. 

Multiple models achieve 0\% accuracy on minority classes: Naive Bayes fails completely on SL, NOL, CRES, BREMOD, and BERGDUC; all neural models fail on NOL and BREMOD; Deep CNN and Transformer achieve 0\% on several classes. Neural models require substantially more training data to learn distinctive orthographic patterns than traditional approaches.

\textbf{Feature Encoding Comparison}
For traditional models, byte-level features often perform competitively with or better than character-level features on balanced metrics: Logistic Regression with byte features achieves 85.78\% average class accuracy versus 84.96\% for character features and 84.33\% for combined features. Word-level features, when used alone, underperform other encoding types, but contribute to improved performance when combined with character and byte features in traditional models. Word features with logistic regression lead to the best accuracy for BREMOD. Also worth noting, combining all features significantly improves performance on the same classifier for LORUNIF.

For neural models, the choice between byte and character encoding has minimal impact on overall performance (CNN: 94.27\% vs 94.18\%, LSTM: 90.60\% vs 91.05\%), though character-level encoding shows slight advantages for some classes. 

\textbf{Confusion Analysis}
We identify systematic confusion patterns common across all models. The most frequent error is misclassifying LOCC as MILCLASS (56 total errors across all models), which is unsurprising given both are Western Lombard orthographies with overlapping graphemes and features. The reverse confusion occurs less frequently (24 times).

Eastern Lombard orthographies also show expected mutual confusion. BREMOD is misclassified as LORUNIF (13 errors). The pan-Lombard orthographies (SL, NOL) are misclassified as MILCLASS when errors occur. This pattern may result from multiple factors: insufficient training examples to distinguish their distinctive features; lexical influence, as many SL and NOL articles are likely written by Western Lombard speakers; and the fact that both orthographies are fundamentally based on MILCLASS, with SL being more distinctive, while NOL remains closer to the classical milanese standard. This is also exacerbated by the fact that models are incentivized to classify dubious examples as MILCLASS, being it the most frequent class in the training data. 

Looking at the output of the \texttt{Log.byte} model, qualitative analysis on classification errors regarding Eastern Lombard orthographies (the hardest ones for the model) shows that confusion arises due to the high degree of similarity between BERGDUC, BREMOD, and LORUNIF. In many cases of LORUNIF false positives, the text contained named entities, such as toponyms, or not enough orthographical information to allow for meaningful distinction. On the other hand, when LORUNIF is classified as either BERGDUC or BREMOD, the model may be influenced by lexical choices. This may be also the reason behind the almost perfect performance on CRES for logistic regression and SVM models. While on the surface CRES is very close to BERGDUC, BREMOD, and LORUNIF, lexical differences (e.g. \textit{al} vs BERGDUC \textit{ol} vs BREMOD \textit{el}; masculine singular "the") may drive the models to tag it correctly.

% NO-TAG RESULTS with LOG. BYTES
We apply the same \texttt{Log.byte} model on the 94,520 untagged and unfiltered Wikipedia lines. Some of the patterns already discussed are observable also here. The resulting distribution is heavily skewed towards Western Lombard, with MILCLASS and LOCC being 81.0\% of the data. The overall prediction confidence is relatively low, with an average score of 0.33. This is due to the brevity of the untagged samples: short sequences, such as foreign place names, lack the necessary orthographic and linguistic features to allow for high-confidence classification. This drives the model to assign them to the most frequent MILCLASS tag.

\section{Conclusions}

This paper presented LombardoGraphia, a curated corpus for the Lombard language, tagged by its main orthography variants. We used this dataset to experiment with and train traditional and neural models to automatically detect the variety of a given Lombard text. We make available both of these resources, to further NLP research for the Lombard language.

While our models achieve near perfect accuracy on majority classes, and common confusion patterns are still between very similar orthographies, improving the performance on minority classes is also desirable, if our models were to be used to build datasets for further research.

Thus, some work is still left for the future: a wider dataset to increase the coverage of minority classes and the models' accuracy for their classification; testing a document-level approach to classification; and apply these models to digitalized books and other sources, possibly with different orthographic variants. The classification itself can be improved by experimenting with other approaches, such as clustering and multi-label classification, and by adding an option for non-Lombard text.

These issues not withstanding, we believe that the dataset and the models in this work represent a critical stepping stone toward variety-aware NLP for Lombard.

\section*{Limitations}

Our work has some limitations. First, our dataset is relatively small and imbalanced. While this is sufficient for most models to correctly classify the majority classes, minority ones are still very hard, especially for neural models. Second, we are limited to study readily available data from Wikipedia. Even if these data are convenient and with reasonably good coverage, they exclude text that could be found in everyday life and published books, most often written in subjective orthographies, which can be both very similar or completely different from the ones used on Wikipedia.

\section*{Ethical Considerations}

NLP, especially when underresourced and endangered languages are involved, should be primarily concerned to work for and with the speakers. With this work, we hope to provide Lombard speakers and language activists with a useful tool that can further the presence and vitality of the Lombard language. 

We are aware that our models are not perfect, especially when dealing with underrepresented orthographic variants, and thus their predictions should be taken with reasonable caution.  

\section*{Acknowledgements}

We would like to acknowledge the volunteers contributing to and maintaining the Lombard Wikipedia without whom these data would not exist. We would like to thank the reviewers for their useful comments. This work has been partly supported by the Ministry of Education, Youth and Sports of the Czech Republic within the LINDAT-CLARIAH-CZ project LM2023062.

\section*{Bibliographical References}

\bibliographystyle{lrec2026-natbib}
%\bibliography{lrec2026-example,anthology,anthology_2}

\section*{Language Resource References}

\bibliographystylelanguageresource{lrec2026-natbib}
%\bibliographylanguageresource{languageresource}

\end{document}